%% file: main.tex
\documentclass[runningheads]{llncs}
\usepackage{graphicx}
\usepackage[colorlinks,linkcolor=red, anchorcolor=blue, citecolor=green]{hyperref}
\usepackage{tikz}
\usepackage{comment}
\usepackage{amsmath,amssymb} 
\usepackage{color}
\usepackage{multirow}
\usepackage{wrapfig}
\usepackage[accsupp]{axessibility}  

\newcommand{\ie}{\textit{i}.\textit{e}.}
\newcommand{\eg}{e.g.}
\newcommand{\vs}{\textit{vs}. }
\newcommand{\trans}{\emph{transformer}}
\usepackage[flushleft]{threeparttable}

\makeatletter
\renewcommand\normalsize{%
\@setfontsize\normalsize\@xpt\@xiipt
\abovedisplayskip 2\p@ \@plus2\p@ \@minus5\p@
\abovedisplayshortskip \z@ \@plus3\p@
\belowdisplayshortskip 6\p@ \@plus3\p@ \@minus3\p@
\belowdisplayskip \abovedisplayskip
\let\@listi\@listI}
\makeatother

\begin{document}
\pagestyle{headings}
\mainmatter
\def\ECCVSubNumber{1735}  

\title{Video Graph Transformer for Video Question Answering} 

%

\author{
Junbin Xiao\inst{1,2,3} \and
Pan Zhou\inst{1} \and
Tat Seng Chua \inst{2,3} \and
Shuicheng Yan\inst{1}
}
\authorrunning{Xiao et al.}
\institute{Sea AI Lab \and Sea-NExT Joint Lab, Singapore \and
Department of Computer Science, National University of Singapore \\
\email{junbin@comp.nus.edu.sg, zhoupan@sea.com, dcscts@nus.edu.sg, yansc@sea.com}
}

\maketitle

\begin{abstract}
This paper proposes a Video Graph Transformer (VGT) model for Video Quetion Answering (VideoQA).
VGT's uniqueness are two-fold: 1) it designs a dynamic graph transformer module which encodes video by explicitly capturing the visual objects, their relations, and dynamics for complex spatio-temporal reasoning; and 2) it exploits disentangled video and text Transformers for relevance comparison between the video and text to perform QA, instead of entangled cross-modal Transformer for answer classification. Vision-text communication is done by additional cross-modal interaction modules. With more reasonable video encoding and QA solution, we show that VGT can achieve much better performances on VideoQA tasks that challenge dynamic relation reasoning than prior arts in the pretraining-free scenario. Its performances even surpass those models that are pretrained with millions of external data. We further show that VGT can also benefit a lot from self-supervised cross-modal pretraining, yet with orders of magnitude smaller data. These results clearly demonstrate the effectiveness and superiority of VGT, and reveal its potential for more data-efficient pretraining. With comprehensive analyses and some heuristic observations, we hope that VGT can promote VQA research beyond coarse recognition/description towards fine-grained relation reasoning in realistic videos. Our code is available at \url{https://github.com/sail-sg/VGT}.

\end{abstract}

\section{Introduction}
\label{introduction}
Since the 1960s, the very beginning of Artificial Intelligence (AI), long efforts and steady progresses have been made towards machine systems that can demonstrate their understanding of the dynamic visual world by responding to humans' natural language queries in the context of videos which directly reflect our physical surroundings. In particular, since 2019 \cite{devlin2018bert}, we have been witnessing a drastic advancement in such multi-disciplinary AI where computer vision, natural language processing as well as knowledge reasoning are coordinated for accurate decision making. This advancement stems, in part from the success of \emph{multi-modal pretraining} on web-scale vision-text data \cite{chen2020uniter,jia2021scaling,li2021align,li2020oscar,lu2019vilbert,radford2021learning,su2020vl,sun2019videobert,tan2019lxmert,xu2021videoclip}, and in part from the unified deep neural network that can well model both vision and natural language data, \ie, \trans~\cite{vaswani2017attention}. As a typical multi-disciplinary AI task, Video Question Answering (VideoQA) has benefited a lot from these developments which helps to propel the field steadily forward over the use of purely conventional techniques \cite{fan2019heterogeneous,gao2018motion,jang2017tgif,jiang2020reasoning,le2020hierarchical,xiao2021video,zhong2022video}.
\vspace{-0.1cm}

Despite the excitement, we find that the advances made by such \trans-style models mostly lie in answering questions that demand the holistic recognition or description of video contents \cite{lei2018tvqa,seo2021look,xu2017video,xu2021videoclip,yang2021just,yu2018joint,zhu2020actbert}. The problem of answering questions that challenge real-world visual relation reasoning, especially the causal and temporal relations that feature video dynamics \cite{jang2017tgif,xiao2021next}, is largely under-explored. Cross-modal pretraining seems promising \cite{lei2021less,yu2021learning,zellers2021merlot}. Yet, it requires the handling of prohibitively large-scale \emph{video}-text data \cite{fu2021violet,zellers2021merlot}, or otherwise the performances are still inferior to the state-of-the-art (SoTA) conventional techniques \cite{lei2021less,seo2021attend,yu2021learning}. In this work, we reveal two major reasons accounting for the failure: 
1) \textbf{Video encoders are overly simplistic.} Current video encoders are either 2D neural networks (CNNs~\cite{he2016deep,ren2015faster} or Transformers~\cite{dosovitskiy2020image}) operated over sparse frames or 3D neural networks ~\cite{bertasius2021space,liu2021video,xie2018rethinking} operated over short video segments. Such networks encode the videos holistically, but fail to explicitly model the fine-grained details, \ie, spatio-temporal interactions between visual objects. Consequently, the resulting VideoQA models are weak in reasoning and require large-scale video data for learning to compensate for such weak forms of input.
2) \textbf{Formulation of VideoQA problem is sub-optimal}. Often, in multi-choice QA, the video, question, and each candidate answer are appended (or fused) into one holistic token sequence and fed to a cross-modal Transformer to gain a global representation for answer classification \cite{zhu2020actbert,lei2021less}. Such a global representation is weak in disambiguating the candidate answers, because the video and question portions are the same and large, which may overwhelm the short answer and dominate the overall representation. In open-ended QA (popularly formulated as a multi-class classification problem \cite{xu2017video}), answers are treated as class indexes and their word semantics (which are helpful for QA.) are ignored. The insufficient information modelling exacerbates the data-hungry issue and leads to sub-optimal performance as well.
\vspace{-0.1cm}

To improve visual relation reasoning and also reduce the data demands for video question answering, we propose the Video Graph Transformer (VGT) model. VGT addresses the aforementioned problems and advances over previous \trans-style VideoQA models mainly in two aspects:
1) For video encoder, it designs a dynamic graph transformer module which explicitly captures the objects and relations as well as their dynamics to improve visual reasoning in dynamic scenario. 
2) For problem formulation, it exploit \emph{separate} vision and text transformers to encode video and text respectively for similarity (or relevance) comparison instead of using a single cross-modal transformer to fuse the vision and text information for answer classification. Vision-text communication is done by additional cross-modal interaction modules.
Through more sufficient video information modelling and more reasonable QA problem solution, we show that VGT can achieve much better performances on benchmarks featuring dynamic relation reasoning than previous arts including those pretrained on million-scale vision-text data. Such strong performance comes even without using external data to pretrain. When pretraining VGT with a small amount of data, we can observe further and non-trivial performance improvements. The results clearly demonstrate VGT's effectiveness and superiority in visual reasoning, as well as its potential for more data-efficient\footnote{The model demands on less training data to achieve good performance.} video-language pretraining.

To summarize our contributions: 1) We propose Video Graph Transformer (VGT) that advances VideoQA from shallow description to in-depth reason. 2) We design a dynamic graph transformer module which shows strength for visual reasoning. The module is task-agnostic and can be easily applied to other video-language tasks. 3) We achieve SoTA results on NExT-QA \cite{xiao2021next} and TGIF-QA \cite{jang2017tgif} that task visual reasoning of dynamic visual contents. Also, our structured video representation gives a promise for data-efficient video-language pretraining.

\section{Related Work}
\vspace{-0.3cm}
\textbf{Conventional Techniques for VideoQA.}
Prior to the success of Transformer for vision-language tasks, various techniques, \eg, cross-modal attention \cite{jang2017tgif,li2019beyond,jiang2020divide}, motion-appearance memory \cite{gao2018motion,fan2019heterogeneous,liu2021hair}, and graph neural networks \cite{jiang2020reasoning,li2022invariant,park2021bridge}, have been proposed to model informative videos contents for answering questions. Yet, most of them leverage frame- or clip-level video representations as information source.  Recently, graphs constructed over object-level representations \cite{huang2020location,liu2021hair,seo2021attend,xiao2021video} have demonstrated superior performance, especially on benchmarks that emphasize visual relation reasoning \cite{jang2017tgif,shang2019annotating,shang2019relation,xiao2021next}.
However, these graph  methods either construct monolithic graphs that do not disambiguate between relations in 1) space and time, 2) local and global scopes \cite{huang2020location,wang2018videos}, or build static graphs at frame-level without explicitly capturing the temporal dynamics \cite{liu2021hair,peng2021progressive,xiao2021video}. The monolithic graph is cumbersome to long videos where multiple objects interact in space-time. Besides, the static graphs may lead to incorrect relations (\eg, \texttt{hug} \vs \texttt{fight}) or fail to capture dynamic relations (\eg, \texttt{take away}). In this work, we model video as a local-to-global dynamic visual graph, and design graph transformer module to explicitly model the objects, their relations, and dynamics, for  exploiting object and relations in adjacent frames to calibrate the spurious relations obtained at static frame-level. Importantly, we also integrate strong language models and explore cross-modal pretraining techniques to learn the structured video representations in a self-supervised manner.

\textbf{Transformer for VideoQA.}
Pioneer works \cite{li2020hero,seo2021look,xu2021videoclip,yang2021just,zhu2020actbert} learn generalizable representations from HowTo100M \cite{miech2019howto100m} by either applying various proxy tasks \cite{zhu2020actbert}, or curating more tailored-made supervisions (\eg, future utterance \cite{seo2021look} and QA pairs \cite{yang2021just}) for VideoQA. However, they focus on answering questions that demand the holistic recognition \cite{xu2017video} or shallow description \cite{yu2018joint}, and their performances on visual relation reasoning \cite{jang2017tgif,xiao2021next} remains unknown. Furthermore, recent works \cite{bain2021frozen,zellers2021merlot} reveal that these models may suffer from performance lose on open-domain questions due to the heavy noise \cite{amrani2021noise,miech2020end} and limited data scope 
of HowTo100M.
Recent efforts tend to use open-domain vision-text data for end-to-end learning. ClipBERT \cite{lei2021less} takes advantage of image-caption data \cite{chen2015microsoft,krishna2017visual} for pretraining, but it only has limited performance improvement on temporal reasoning tasks \cite{jang2017tgif}, as the temporal relations are hard to learn from static images. In addition, ClipBERT relies on human annotated descriptions which are expensive to annotate and hard to scale up. More recent works \cite{fu2021violet,zellers2021merlot} collect million-scale user-generated (vastly abundant on the Web) vision-text data \cite{bain2021frozen,sharma2018conceptual,zellers2021merlot} for pretraining, but suffers from huge computational cost to train on such large-scale datasets. Two latest works~\cite{buch2022revisiting,ding2021attention} reveal the potential of Transformers for learning on the target datasets (relatively small scale). While promising, they either target at revealing the single-frame bias of benchmark datasets by using image-text pretrained features (\eg~from CLIP \cite{radford2021learning}), or only demonstrate the model's effectiveness on synthesized data \cite{yi2019clevrer}.
Overall, the \emph{poor-dynamic-reasoning} and \emph{data-hungry} problems in existing \trans-style video-language models largely motivate this work. To alleviate these problems, we explicitly model the objects and relations for dynamic visual reasoning and incorporate structure priors (or relational inductive bias \cite{battaglia2018relational}) into transformer architectures to reduce the demand on data.

\textbf{Graph Transformer.}
The connection between graph neural networks and Transformer has earned increasing attention \cite{wang2021tcl,ying2021transformers,yun2019graph}. Nonetheless, the major advancements are made in modelling natural graph data (\eg~social connections) by either incorporating graph expertise (\eg, node degrees) into self-attention block of Transformer \cite{ying2021transformers}, or designing \trans-style convolution blocks to fuse information from heterogeneous graphs \cite{yun2019graph}. A recent work \cite{geng2021dynamic} combines graphs and Transformers for video dialogues. Yet, it simply applies global transformer over pooled graph representations built from static frames and does not explicitly encode object and relation dynamics. Our work differs from it by designing and learning dynamic visual graph over video objects and using transformers to capture the temporal dynamics at both local and global scopes.

\vspace{-0.2cm}
\section{Method}
\vspace{-0.2cm}
\subsection{Overview}
Given a video \textit{v} and a question \textit{q}, VideoQA aims to combine the two stream information \textit{v} and \textit{q} to predict the answer \textit{a}. Depending on the task settings, \textit{a} can be given in multiple choices along with each question for multi-choice QA, or it is given in a global answer set for open-ended QA. In this work, we handle both types of VideoQA by optimizing the following objective:
\begin{equation}\label{Obj}
    a^* = \arg\max\nolimits_{a \in \mathcal{A}} {\mathcal{F}_W(a | q, v, \mathcal{A})}, 
\end{equation}
in which $\mathcal{A}$ can be $\mathcal{A}_{mc}$ corresponding to the candidate answers of each question in multi-choice QA, or $\mathcal{A}_{oe}$ corresponding to the global answer set in open-ended QA. $\mathcal{F}_W$ denotes the mapping function with learnable parameters $W$.

\begin{wrapfigure}[11]{r}{0.5\textwidth}
\vspace{-12pt}
  \begin{center}
    \includegraphics[width=.5\textwidth]{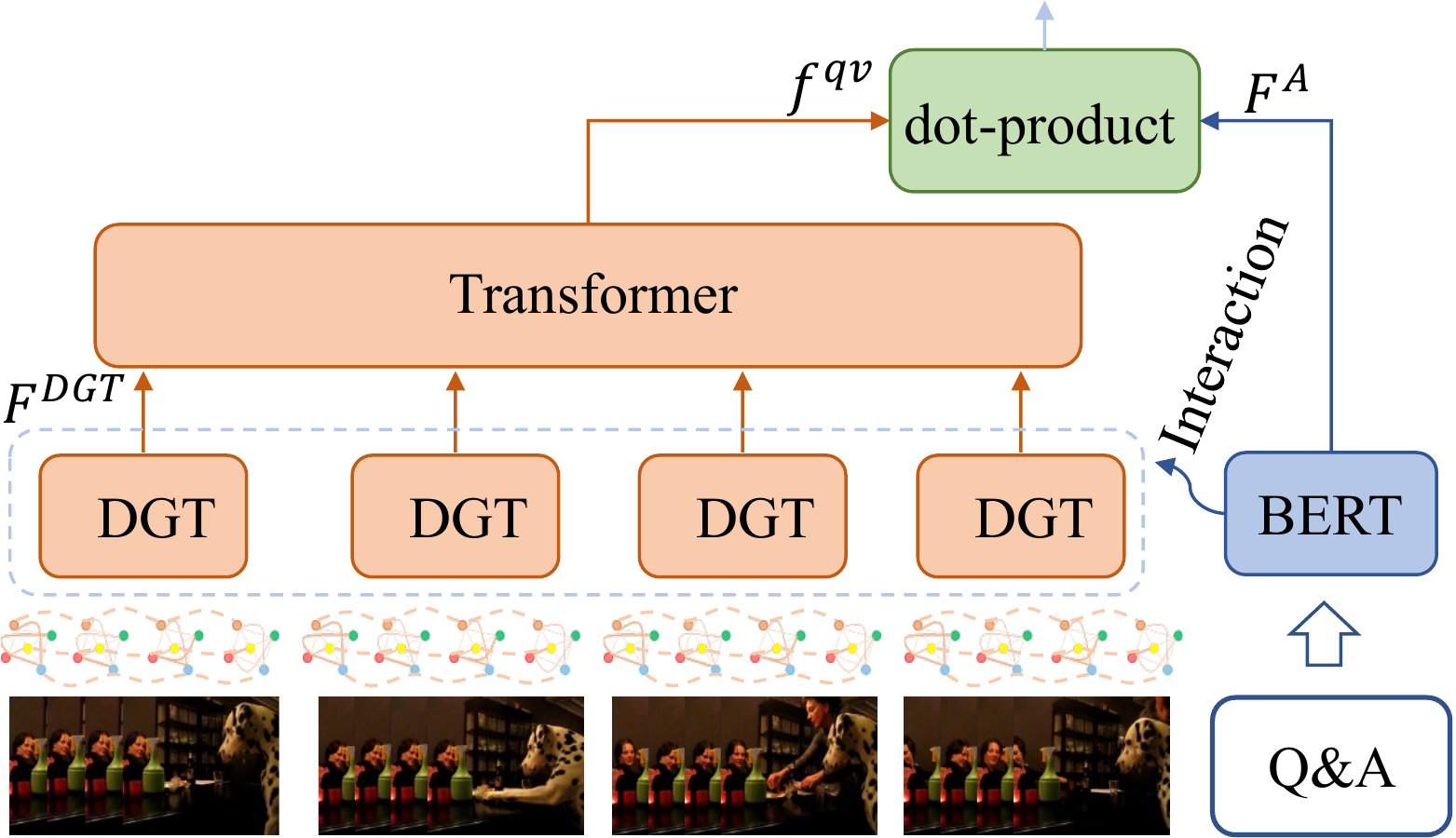}
  \end{center}
\vspace{-20pt}
  \caption{Overview of video graph transformer (VGT) for VideoQA.}
  \label{fig:framework}
\end{wrapfigure}

To solve the problem, we design a video graph transformer (VGT) model to perform the mapping $\mathcal{F}_W$ in Eqn.~\eqref{Obj}. As illustrated in Fig.~\ref{fig:framework}, at the visual part (Orange), VGT takes as input visual object graphs, and drives a global feature $f^{qv}$ with the integration of textual information, to represent the query-relevant video content. At the textual part (Blue), VGT extracts the feature 
representations $F^\mathcal{A}$  for all the candidate answers via a language model (\eg, BERT~\cite{devlin2018bert}). The final answer $a^*$ is determined by returning the candidate answers with maximal similarity (relevance score) between $f^{qv}$ and $f^a \in F^\mathcal{A}$ via dot-product. At the heart of the model is the dynamic graph transformer module (DGT). The module clip-wisely reasons over the input graphs, and aggregates them into a sequence of feature representations $F^{\text{DGT}}$ which are then fed to a global transformer to achieve $f^{qv}$. 
During training, the whole framework is end-to-end optimized with Softmax cross-entropy loss. For pretraining with weakly-paired video-text data, we adopt cross-modal matching as the major proxy task and optimize the model in a contrastive manner \cite{radford2021learning} along with masked language modelling \cite{devlin2018bert}.
 
\begin{wrapfigure}[11]{r}{0.5\textwidth}
\vspace{-32pt}
  \begin{center}
    \includegraphics[width=.5\textwidth]{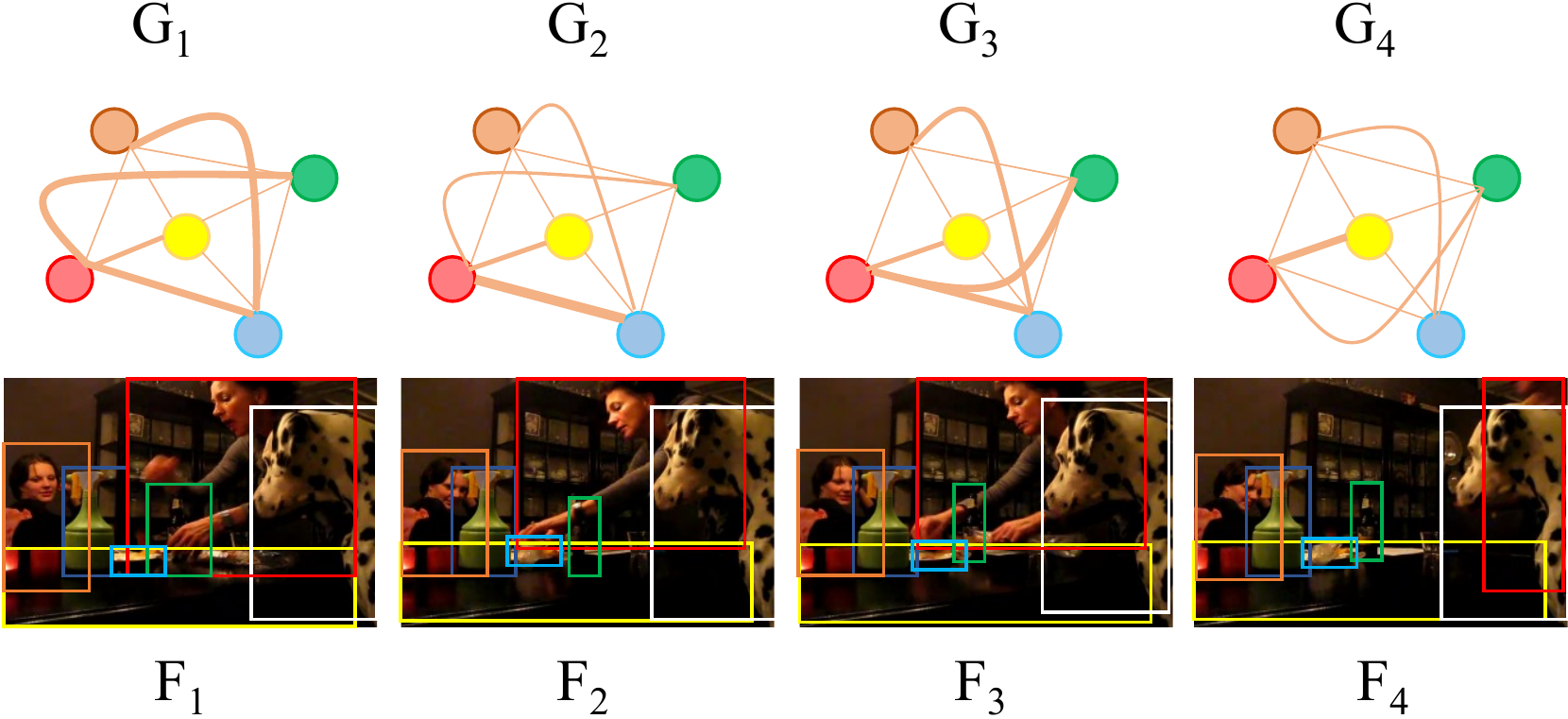}
  \end{center}
\vspace{-20pt}
  \caption{Illustration of graph construction in a short video clip of $l_c=4$ frames. The nodes of same color denote same object.}
  \label{fig:vgr}
\end{wrapfigure}
\subsection{Video Graph Representation}
\label{sec:vgr}
Given a video, we sparsely sample $l_v$ frames in a way analogous to \cite{xiao2021video}. The $l_v$ frames are evenly distributed into $k$ clips of length $l_c=\frac{l_v}{k}$.
For each sampled frame (see Fig.~\ref{fig:vgr}), we extract $n$ RoI-aligned features as object appearance representations $F_r\!=\!\{f_{r_i}\}_{i=1}^n$ along with their spatial locations $B\!=\!\{b_{r_i}\}_{i=1}^n$ with a pretrained object detector~\cite{anderson2018bottom,ren2015faster}, where $r_i$ represents the $i$-th object region in a frame. Additionally, we obtain an image-level feature $F_I\!=\!\{f_{I_t}\}_{t=1}^{l_v}$ for all the sampled frames with a pretrained image classification model~\cite{he2016deep}. $F_I$ serve as global contexts to augment the graph representations aggregated from the local objects. 

To find the same object across different frames within a clip, we define a linking score $s$ by considering their appearance and spatial location:
\begin{equation}
    s_{i, j} =  \psi(f_{r_i}^t, f_{r_j}^{t+1}) + \lambda*\text{IoU}(b_i^t, b_j^{t+1}), \quad t\in\{1, 2, \dots, l_c-1\},
\end{equation}
where $\psi$ denotes the cosine similarity between two detected objects $i$ and $j$ in adjacent frames.  Intersection-over-union (IoU) computes the location overlap of objects $i$ and $j$. Our experiments always set  $\lambda$ as one. The $n$ detected objects in the first frame of each clip are designated as anchor objects. Detected objects in consecutive frames are then linked to the anchor objects by greedily maximizing $s$ frame by frame\footnote{We assume that the group of objects do not change in a short video clip.}. By aligning objects within a clip, we ensure the consistency of the node and edge representations for the graphs constructed at different frames.

Next, we concatenate the object appearance $f_r$ and location $f_{loc}$ representations and project the combined feature into the $d$-dimensional space via
\begin{equation}
\label{equ:fo}
    f_o = \mathrm{ELU}(\phi_{W_o}([f_r;f_{loc}])),
\end{equation}
where $[;]$ denotes feature concatenation and $f_{loc}$ is obtained by applying a $1 \times 1$ convolution over the relative coordinates as in \cite{xiao2021video}. The function $\phi_{W_o}$ denotes a linear transformation with parameters $W_o$. With $F_o\!=\!\{f_{o_i}\}_{i=1}^n$, the relations in the $t$-th frame can be initialized as 
pairwise similarities:
\begin{equation}
\label{equ:aa}
    R_t = \sigma(\phi_{W_{ak}}(F_{o_t})\phi_{W_{av}}(F_{o_t})^\mathrm{\top}), \quad t\in\{1, 2, \dots, l_v\},
\end{equation}
where $\phi_{W_{ak}}$ and $\phi_{W_{av}}$ denote linear transformations with parameters $W_{ak}$ and $W_{av} \in \mathbb{R}^{d \times \frac{d}{2}}$ respectively. We use different transformations to reflect the asymmetric nature of real-world subject-object interactions \cite{krishna2018referring,xiao2020visual}. For symmetric relations, we expect that the learned parameters $W_{ak}$ and $W_{av}$ are quite similar. $\sigma$ is the Softmax operation that normalizes each row. 
For brevity, we use $G_t=(F_{o_t}, R_t)$ to denote the graph representation of the $t$-th frame where $F_{o}$ are node representations and $R$ are edge representations of the graph.
\vspace{-0.3cm}

\subsection{Dynamic Graph Transformer}
Our dynamic graph transformer (DGT)  takes as input a set of visual graphs $\{G_t\}_{t=1}^{L_v}$ clip-wisely, and outputs a sequence of representations $F^{DGT}\in\mathbb{R}^{d\times k}$ by mining the temporal dynamics of objects and their spatial interactions. To  this end, we sequentially operate    a temporal graph transformer unit, a spatial graph convolution unit and a hierarchical aggregation unit as detailed below.
\vspace{-0.3cm}
\subsubsection{Temporal Graph Transformer}
\begin{wrapfigure}[12]{r}{0.5\textwidth}
 \vspace{-10pt}
  \begin{center}
    \includegraphics[width=.5\textwidth]{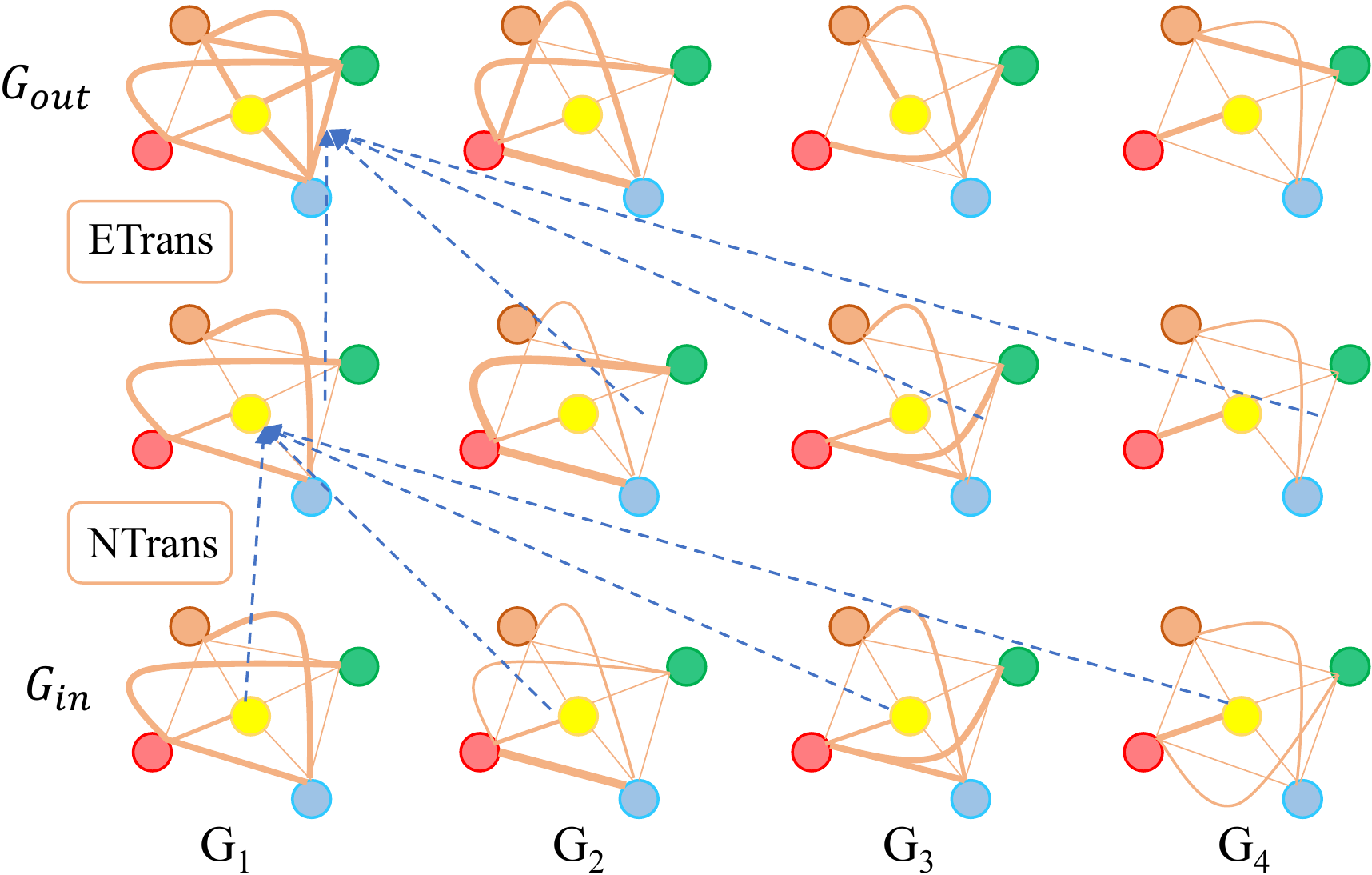}
  \end{center}
   \vspace{-20pt}
  \caption{Illustration of temporal graph transformer in a short video clip.}
  \label{fig:tgt}
\end{wrapfigure}

As illustrated in Fig.~\ref{fig:tgt}, the temporal graph transformer unit takes as input a set of graphs $G_{in}$ and outputs a new 
set of graphs $G_{out}$ by mining the temporal dynamics among them via a node transformer (NTrans) and an edge transformer (ETrans). 
For completeness, we briefly recap the self-attention in Transformer \cite{vaswani2017attention}. It uses a multi-head self-attention (MHSA) to fuse a sequence of input features $X_{in}=\{x_{in}^t\}_{t=1}^l$:
\begin{equation}
X_{out} = \text{MHSA}(X_{in}) = \phi_{W_c}([h_1;h_2;\dots,h_e]),
\end{equation}
where $\phi_{W_c}$ is a linear transformation with parameters $W_c$, and
\begin{equation}
    h_i = \text{SA}(\phi_{W_{i_q}}(X_{\text{in}}), \phi_{W_{i_k}}(X_{\text{in}}), \phi_{W_{i_v}}(X_{\text{in}})),
\end{equation}
where $\phi_{W_{i_q}}$, $\phi_{W_{i_k}}$ and $\phi_{W_{i_v}}$ denote the linear transformations of the query, key, and value vectors of the $i$-th self-attention (SA) head respectively. $e$ denotes the number of self-attention heads, and SA is defined as:
\begin{equation}
    \text{SA}(X_q, X_k, X_v) = \sigma\left({X_k X_q^\mathrm{\top}/\sqrt{d_k}}\right)X_v, 
\end{equation}
in which $d_k$ is the dimension of the key vector. Finally, a skip-connection with layer normalization (LN) is applied to the output sequence $X=LN(X_{out}+X_{in})$. $X$ can undergo more MHSAs depending on the number of transformer layers.

In  temporal graph transformer, we apply $H$ self-attention blocks to enhance the node (or object) representations by aggregating information from other nodes of the same object from all adjacent frames within a clip:
\begin{equation}
\label{equ:node}
    F'_{o_i} = \text{NTrans}(F_{o_i})=\text{MHSA}^{(H)}(F_{o_i}),
\end{equation}
in which $F_{o_i} \in \mathbb{R}^{l_c \times d}$ denotes a sequence of feature representations corresponding to object $i$ in a video clip of length $l_c$. Our motivation behind the node
transformer is that it models the change of single object behaviours and thus infer the dynamic actions (\eg~\texttt{bend down}). Also, it is helpful in improving the objects' appearance feature in the cases where the object at certain frames suffer from motion blur or partial occlusion.

Based on the new nodes $F'_o=\{F'_{o_i}\}_{i=1}^n$, we  update the relation matrix $R$ via Eqn.~\eqref{equ:aa}. Then, to explicitly model the temporal relation dynamics, we apply an edge transformer on the updated relation matrices: 
\begin{equation}
\label{equ:edge}
    \mathcal{R'} =\text{ETrans}(\mathcal{R}) =\text{MHSA}^{(H)}(\mathcal{R}),
\end{equation}
where $\mathcal{R}\!=\!\{R_t\}_{t=1}^l\in\mathbb{R}^{l_c \times d_n}$ ($d_n=n^2$) is the $l_c$ adjacency matrices that are row-wisely expanded. Our motivation is that the relations captured at static frames may be spurious, trivial or incomplete. The edge transformer can help to calibrate the wrong relations and recall the missing ones. 
For brevity, we refer to the temporally contextualized graph at the $t$-th frame as $G_{out_t}=(F'_{o_t}, R'_t)$. 
\vspace{-0.3cm}

\subsubsection{Spatial Graph Convolution}
The temporal graph transformer focuses on temporal relation reasoning. To reason over the object spatial interactions, we apply a $U$-layer graph attention convolution \cite{kipf2016semi} on all the $l_v$ graphs:
\begin{equation}
\label{equ:gcn}
    {F'_o}^{(u)} = \text{ReLU}((R'+I){F'_o}^{(u-1)}W^{(u)}),
\end{equation}
where $W^{(u)}$ is the graph parameters at the $u$-th layer. $I$ is the identity matrix for skip connections. ${F'_o}^{(u)}$ are initialized by  the output node representations $F'_o$ as aforementioned. The index $t$ is omitted for brevity. A last skip-connection: $F_{o_{out}}=F'_o+{F'_o}^{(U)}$ is used to obtain the final node representations.

\vspace{-0.3cm}
\subsubsection{Hierarchical Aggregation}
\begin{wrapfigure}[7]{r}{0.4\textwidth}
 \vspace{-32pt}
  \begin{center}
    \includegraphics[width=.4\textwidth]{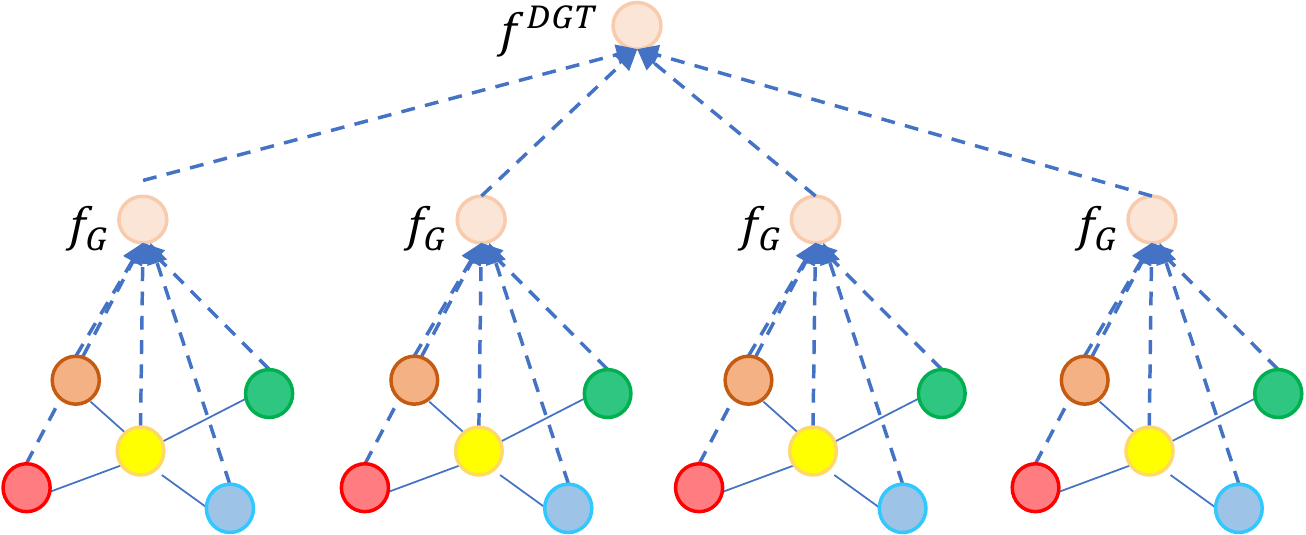}
  \end{center}
   \vspace{-20pt}
  \caption{Hierarchical Aggregation.}
  \label{fig:hpool}
\end{wrapfigure}

The node representations so far have explicitly token into account the objects' spatial and temporal interactions. But 
such interactions are mostly atomic. To aggregate these
atomic interactions into higher-level
video elements, we adopt a hierarchical aggregation strategy in Fig.~\ref{fig:hpool}. 
First, we aggregate the graph nodes at each frame by a simple attention:
\begin{equation}
   f_G = \sum\nolimits_{i=1}^N \alpha_i F_{o_{out_i}}, \quad  \alpha = \sigma(\phi_{W_G}(F_{o_{out}})),
\end{equation}
where $\phi_{W_G}$ is linear transformation with parameters $W_G \in \mathbb{R}^{d \times 1}$.
The graph representation $f_G$ captures a local object interactions. It may lose sight of a global picture of a frame, especially 
since we only retain $n$ objects and cannot guarantee that they include all the objects of interest in that frame. As such, we complement $f_G$ with the frame-level feature $f_I$ by concatenation:
\begin{equation}
\label{equ:ff}
    f_G = \text{ELU}(\phi_{W_m}([\phi_{W_f}(f_I);f_G]))
\end{equation}
in which $\phi_{W_m}$ and $\phi_{W_f}$ are linear transformations with parameters $W_m\in\mathbb{R}^{2d \times d}$ and $W_f\in\mathbb{R}^{2048\times d}$ respectively.
We next pool the local interactions to obtain a sequence of clip-level feature representations via:
\begin{equation}
\label{equ:fc}
 f^{\text{DGT}}=\text{MPool}(F_G)=\frac{1}{l_c}\sum\nolimits_{t=1}^{l_v} f_{G_t}
\end{equation}
\noindent The set of $k$ clips are finally represented by  $F^{\text{DGT}}\!=\!\{f_c^{\text{DGT}}\}_{c=1}^k$. 

\subsection{Cross-modal Interaction}
\label{sec:cm}
To find the informative visual contents with respect to a particular text query, a cross-model interaction between the visual and textual nodes is essential. Given a set of visual nodes denoted by $X^v$, we integrate textual information $X^q=\{x_m^q\}_{m=1}^M$   into the visual nodes via a simple cross-modal attention:
\begin{equation}
x^{qv} = x^v + \sum\nolimits_{m=1}^M \beta_m x^q_m, \quad \text{where} \quad \beta = \sigma(x^v(X^q)^\top),
\end{equation}
where $M$ is the number of tokens in the text query. 
In principle, the $X^v$ can be visual representations from different levels of the DGT module similar to \cite{xiao2021video}. In our experiment, we explore performing the cross-modal interaction with visual representations at the object-level ($F_O$ in Eqn.~\eqref{equ:fo}), frame-level ($F_G$ in Eqn.~\eqref{equ:ff}), and  clip-level ($F^{DGT}$ in Eqn.~\eqref{equ:fc}). We find that the results vary among different datasets. As a default, we perform cross-modal interaction at the clip-level outputs (\ie, the outputs of the DGT module $X^v:=F^{\text{DGT}}$), since the number of nodes at this stage is much smaller, and the node representations have already absorbed the information from the preceding layers. For the text node $X^q$, we obtain them by a simple linear projection on the token outputs of a language model \cite{devlin2018bert}: 
\begin{equation}
\label{equ:bproj}
    X^q=\phi_{W_Q}(\text{BERT}(Q)),
\end{equation}
where $W_Q\in\mathbb{R}^{768 \times d}$. The text query Q can be questions in open-end QA or QA pairs in multi-choice QA. Note that in multi-choice QA, we max-pool the obtained query-aware visual representations with respect to different QA pairs to find the one that is mostly relevant to the video. 

\subsection{Global Transformer}
\label{sec:gt}
The aforementioned DGT module pays attention to extract informative visual clues from video clips. To capture the temporal dynamics between these clips, we employ another $H$-layer transformer over the cross-modal interacted clip feature (\ie~$F^{\text{DGT}}$), and add learnable sinusoidal temporal position embeddings~\cite{devlin2018bert}. Finally, the transformer's outputs are mean-pooled to obtain the global representation $f^{qv}\in\mathbb{R}^d$ for the entire video, which is defined as follows:
\begin{equation}
\label{equ:gbtrans}
    f^{qv} = \text{MPool}(\text{MHSA}^{(H)}(F^{\text{DGT}})).
\end{equation}
The global transformer has two major advantages: 1) It retains the overall hierarchical structure which progressively drives the video elements at different granularity as in \cite{xiao2021video}. 2) It improves the feature compatibility of vision and text, which may benefit cross-modal comparison.

\subsection{Answer Prediction}
\label{sec:ad}
To obtain a global representation for a particular answer candidate, we mean-pool its token representations from BERT by 
$
    f^A = \text{MPool}(X^A),
$
where $X^A$ denotes a candidate answer's token representations, and is obtained in a way analogous to Eqn.~\eqref{equ:bproj}. Its similarity with the query-aware video representation $f^{qv}$ is then obtained via a dot-product. Consequently, the candidate answer of maximal similarity is returned as the final prediction:
\begin{equation}
    \quad s=f^{qv}(F^A)^\mathcal{\top}, \quad a^* = \arg \max(s),
\end{equation}
in which $F^A=\{f^A_a\}_{a=1}^{|\mathcal{A}|}\in\mathbb{R}^{|\mathcal{A}|\times d}$, and $|\mathcal{A}|$ denotes the number of candidate answers.
Additionally, for open-ended QA, we follow previous works \cite{xiao2021video} and enable a video-absent QA by directly computing the similarities between the question representation $f^q$ (obtained in a way similar to $f^A$) and the answer representations $F^A$. As a result, the final answer can be a joint decision: 
\begin{equation}
\label{equ:oe}
    s=f^{qv}(F^A)^\mathcal{\top} \odot f^q(F^A)^{\mathcal{\top}}
\end{equation}
in which $\odot$ is element-wise product. 
During training, we maximize the $\langle$VQ, A$\rangle$ similarity corresponding to the correct answer of a given sample by optimizing the Softmax cross entropy loss function.
$
   \mathcal{L}= -\sum\nolimits_{i=1}^{|\mathcal{A}|} y_i \log s_i,
$
where $s_i$ is the matching score for the $i$-th sample. $y_i=1$ if the answer index corresponds to the $i$-th sample's ground-truth answer and 0 otherwise.

\subsection{Pretraining with Weakly-Paired Data}
\label{sec:pt}
For cross-model matching, we encourage the representation of each video-text interacted representation $f^{qv}$ to be closer to that of its paired description $f^q$ and be far away from that of negative descriptions which are randomly collected from other video-text pairs in each training iteration. This is formally achieved by maximizing the following contrastive objective:
\begin{equation}
    \sum\limits_i \log(\frac{\exp{(f^{qv}_i(f^q_i)^{\top})}}{\exp{(f^{qv}_i(f^q_i)^{\top})}+\sum\nolimits_{(f^{qv}, f^q)\in \mathcal{N}_i} \exp{(f^{qv}(f^q)^{\top})} }),
\end{equation}
where $\mathcal{N}_i$ denotes the representations of all the negative video-description pairs of the $i$-th sample. The parameters to be optimized are hidden in the process of calculating $f^{qv}$ and $f^q$ as introduced above. For negative sampling, we sample them from the whole training set at each iteration. For masked language modelling, we only corrupt the positive description of each video for efficiency.

\section{Experiment}
\subsection{Dataset and Configuration}
We conduct experiments on benchmarks whose QAs feature temporal dynamics: 1) NExT-QA \cite{xiao2021next} is a manually annotated dataset that features causal and temporal object interaction in space-time. 2) TGIF-QA \cite{jang2017tgif} features short GIFs; it asks questions about repeated action recognition, temporal state transition and frame QA which invokes a certain frame for answer. For better comparison, we also experiment on MSRVTT-QA \cite{xu2017video} which challenges a holistic visual recognition or description. Other data statistics are presented in Appendix \ref{app:dset}.

We decode the video into frames following \cite{xiao2021video}, and then sparsely sample $l_v=32$ frames from each video. The frames are distributed into $k=8$ clips whose length $l_c=4$.
For each frame, we detect and keep $N=20$ regions of high confidence for NExT-QA (Top-5 are used in the pretraining-free experiments, refer to our analysis in Appendix~\ref{app:sap}
), and $N=10$ for the other datasets, using the object detection model provided by \cite{anderson2018bottom}. The dimension of the models' hidden states is $d=512$. The default number of layers and self-attention heads in transformer are $H=1$ and $e=8$ ($e=5$ for edge transformer in DGT) respectively. Besides, the number of graph layers is $U=2$. For training, 
we use Adam optimizer with initial learning rate $1\times10^{-5}$ of a cosine annealing schedule. The batch size is set to 64, and the maximum epoch varies from 10 to 30 among different datasets. Our pretraining data ($\sim$ $0.18$M) are collected from WebVid \cite{bain2021frozen}. More details are presented in Appendix \ref{app:imp}.

\subsection{Sate-of-the-Art Comparison}
In Table \ref{tab:resnextqa}, we compare VGT with the prior arts on NExT-QA \cite{xiao2021video}. The results show that VGT surpasses the previous SoTAs by clear margins on both the val and test sets, improving the overall accuracy by 1.6\% and 1.9\% respectively. VGT even outperforms a latest work ATP~\cite{buch2022revisiting} which is based on CLIP features~\cite{radford2021learning} (VGT \vs ATP: 55.02\% \vs 54.3\%), and thus sets the new SoTA results. In particular, we note that such strong results come without considering large-scale cross-modal pretraining. When pretraining VGT with (relatively) small amount of data, we can further increase the results to 56.9\% and 55.7\% on NExT-QA val and test sets respectively (refer to our analysis of Table~\ref{tab:pt} in Sec.~\ref{sec:ptft}). 
\setlength{\tabcolsep}{3.2pt}
\begin{table*}[t!]
    \small
    \centering
    \caption{Results on NExT-QA \cite{xiao2021next}. (Acc@C, T, D: Accuracy for Causal, Temporal and Descriptive questions respectively. *: Results reproduced with the official code.)
    }
    \vspace{-0.3cm}
    \scalebox{0.8}{
    \begin{tabular}{l|c|ccc|c|ccc|c}
    \multirow{2}*{Method} & \multirow{2}*{CM-Pretrain} & \multicolumn{4}{c|}{NExT-QA Val} & \multicolumn{4}{c}{NExT-QA Test} \\ \cline{3-10}
    ~ & ~  & Acc@C & Acc@T & Acc@D & Acc@All & Acc@C & Acc@T & Acc@D & Acc@All \\ 
    \hline
    HGA \cite{jiang2020reasoning} & -  & 46.26 & 50.74 & 59.33 & 49.74 & 48.13 & 49.08 & 57.79 & 50.01 \\
    IGV \cite{li2022invariant} & - & - & - & - & - & 48.56 & 51.67 & 59.64 & 51.34 \\
    HQGA \cite{xiao2021video} & -  &48.48 & 51.24 & 61.65 & 51.42 & \underline{49.04} & {\bf52.28} & 59.43 & \underline{51.75} \\
    P3D-G \cite{cherian2022} &- & \underline{51.33} & \underline{52.30} & 62.58 & \underline{53.40} & - & -& -& - \\
    VQA-T* \cite{yang2021just} & -  & 41.66 & 44.11 & 59.97 & 45.30 & 42.05 & 42.75 & 55.87 & 44.54 \\
    VQA-T* \cite{yang2021just} & How2VQA69M &  49.60 & 51.49 & \underline{63.19} & 52.32 & 47.89 & 50.02 & \underline{61.87} & 50.83 \\ 
    \hline
    {\bf VGT (Ours)} & - & {\bf 52.28} & {\bf55.09} & {\bf64.09} & {\bf 55.02} & {\bf 51.62} & \underline{51.94} & {\bf63.65} & {\bf53.68} \\ 
    \end{tabular}
    }
    \label{tab:resnextqa}
    \vspace{-0.5cm}
\end{table*}

Compared with VQA-T~\cite{yang2021just} which also formulates VideoQA as problem of similarity comparison instead of classification, VGT outperforms it almost in all metrics. The strong results could be due to that VGT explicitly models the object interactions and dynamics for visual reasoning, instead of holistically encoding video clips with S3D~\cite{miech2020end,xie2018rethinking}. 
For a better analysis, we further replace the S3D encoder in VQA-T with our DGT module. As shown in Table~\ref{tab:comp} (S3D $\rightarrow$ DGT), our DGT encoder significantly improves VQA-T's result by 4.7\%, in which most of the improvements are from answering reasoning
\setlength{\tabcolsep}{1pt}
\begin{wraptable}[7]{r}{0.5\textwidth}
\vspace{-10pt}
\small
\caption{Detailed comparison with VQA-T \cite{yang2021just}. CMTrans: Cross-Modal Transformer.}\label{tab:comp}
\vspace{4pt}
\scalebox{0.68}{
    \begin{tabular}{l|c|ccc|c}
    \multirow{2}*{Models} & \multirow{2}*{Size (M)} & \multicolumn{4}{c}{NExT-QA Val} \\ \cline{3-6}
    ~ & ~ & Acc@C & Acc@T & Acc@D & Acc@All  \\ 
    \hline
    VQA-T \cite{yang2021just} & 600 & 41.66 & 44.11 & 59.97 & 45.30 \\ 
    S3D$\rightarrow$DGT & 641 & 47.53 & 48.08 & 62.42 & 50.02 \\
    CMTrans$\rightarrow$CM &573 & 42.27 & 44.29 & 58.17 & 45.40 \\
    \hline
    VGT (DistilBERT) & 346 & 50.71 & 51.67 & 66.41 & 53.46 \\
    VGT (BERT) &511 & 52.28 & 55.09 & 64.09 & 55.02 \\ 
    \end{tabular}
    }
\end{wraptable}
type of questions.  
Aside from the DGT module, we encode the candidate answers in the context of the corresponding question with a single language model, whereas VQA-T encodes Q and A independently with two language models \cite{sanh2019distilbert}.  
Our method improves answer encoding with contexts and reduces the model size (or parameters), as shown in Table~\ref{tab:comp} (VGT (DistilBERT)).
Finally, VQA-T adopts cross-modal transformer to fuse the video-question pair, whereas we design light-weight cross-modal interaction module. The module is more parameter efficient but has little impact on the performances (CMTrans$\rightarrow$CM in Table~\ref{tab:comp}).

Compared with other graph based methods \cite{cherian2022,jiang2020reasoning,xiao2021video}, VGT enjoys several advantages: 1) It explicitly model the temporal dynamics of both objects and their interactions. 2) It solves VideoQA by explicit similarity comparison between the video and text instead of classification. 3) It represents both visual and textual data with Transformers which may improve the feature compatibility and benefit cross-modal interaction and comparison \cite{devlin2018bert}. 4) VGT uses much few frames for training and inference (\eg, VGT \vs HQGA \cite{xiao2021video}: 32 \vs 256), which benefits efficiency for video encoding. The detailed analyses are given in Sec.~\ref{sec:analysis}.

In Table~\ref{tab:resqa}, we compare VGT with previous arts on the TGIF-QA and MSRVTT-QA datasets. The results show that VGT performs pretty well on the tasks of repeating action recognition and state transition that feature temporal dynamics, surpassing the previous pretraining-free SoTA results significantly by 10.6\% (VGT \vs MASN \cite{seo2021attend}: 95.0\% \vs 84.4\%) and 6.8\% (VGT \vs MHN \cite{peng2022multilevel}: 97.6\% \vs 90.8\%) respectively. It even beats the pretraining SoTA (\ie~MERLOT \cite{zellers2021merlot}) by about 1.0\%, yet without using external data for cross-modal pretraining. On TGIF-QA-R \cite{peng2021progressive} which is curated by making the negative answers in TGIF-QA more challenging, we can also observe remarkable improvements. Besides, VGT also achieves competitive results on normal descriptive QA tasks as defined in FrameQA and MSRVTT-QA though they are not our focus. 

\setlength{\tabcolsep}{3.3pt}
\begin{table*}[t!]
    \small
    \centering
    \caption{Results on TGIF-QA and MSVTT-QA. $\dagger$ denotes TGIF-QA-R \cite{peng2021progressive} whose multiple choices for repeated action and state transition are more challenging. We grey out the results reported in \cite{peng2021progressive} regarding these two sub-tasks, because the candidate answers are slightly different as we have further rectified the redundant choices.}
     \label{tab:resqa}
    \vspace{-0.2cm}
    \scalebox{0.8}{
        \begin{tabular}{l|c|ccc|cc|c}
        \multirow{2}*{Models} & \multirow{2}*{CM-Pretrain} &  \multicolumn{5}{c|}{TGIF-QA} & MSRVTT \\ \cline{3-7}
        ~ & ~  & Action & Transition & FrameQA & Action$\dagger$ & Transition$\dagger$ & -QA \\ 
        \hline
        LGCN \cite{huang2020location}& - & 74.3 & 81.1 & 56.3 & - & - & -  \\
        HGA \cite{jiang2020reasoning}& -  & 75.4 & 81.0 & 55.1 & - & - & 35.5  \\
        HCRN \cite{le2020hierarchical} & - & 75.0 & 81.4 & 55.9 & \textcolor[rgb]{0.5,0.5,0.5}{55.7} & \textcolor[rgb]{0.5,0.5,0.5}{63.9} & 35.6 \\
        B2A \cite{park2021bridge} & - &75.9 & 82.6 & 57.5 & - & - & 36.9 \\
        HOSTR \cite{dang2021hierarchical} & - &75.0 & 83.0 & 58.0 & - & - & 35.9 \\
        HAIR \cite{liu2021hair} & -  &77.8 & 82.3 & 60.2 & - & - & 36.9  \\
        MASN \cite{seo2021attend} & -  &84.4 & 87.4 & 59.5 &- & - & 35.2 \\
        PGAT \cite{peng2021progressive} & -  &80.6 & 85.7 & 61.1 & \textcolor[rgb]{0.5,0.5,0.5}{\underline{58.7}} & \textcolor[rgb]{0.5,0.5,0.5}{\underline{65.9}} & 38.1 \\
        HQGA \cite{xiao2021video} & - &76.9 & 85.6 & 61.3 &- & - & 38.6  \\ 
        MHN \cite{peng2022multilevel} & - &83.5 & 90.8 & 58.1 &- & - & 38.6  \\ 
        \hline
        ClipBERT \cite{lei2021less} & \text{VG+COCO Caption} &  82.8 & 87.8 & 60.3 & - & -& 37.4  \\
        SiaSRea \cite{yu2021learning} & \text{VG+COCO Caption} &  79.7 & 85.3 & 60.2 & - &- & \underline{41.6} \\
        MERLOT \cite{zellers2021merlot} & \text{Youtube180M, CC3M}  & \underline{94.0} & \underline{96.2} & {\bf69.5} & - & - & {\bf43.1}  \\ \hline
        \textbf{VGT (Ours)} & - & {\bf95.0} & {\bf97.6} & \underline{61.6} & {\bf59.9} & {\bf70.5} & 39.7 \\ 
        \end{tabular}
    }
    \vspace{-0.5cm}
\end{table*}

\subsection{Model Analysis}
\label{sec:analysis}

\textbf{DGT.} 
The middle block of Table~\ref{tab:aba} shows that removing the DGT module (w/o DGT) (i.e. directly summarizing the object representations in each clip) leads to clear performance drops ($\sim$2.0\%) on all tasks that 
challenge spatio-temporal reasoning. We then study the temporal graph transformer module (w/o TTrans) by removing both NTrans and ETrans. It shows better results than removing the whole DGT module. Yet, its performances on tasks featuring temporal dynamics are still weak. We further ablate the temporal graph transformer module to investigate the independent contribution of the node transformer (NTrans) and edge transformer (ETrans). The results (w/o NTrans and w/o ETrans) demonstrate that both transformers
benefit temporal dynamic modelling. Finally, the ablation study on the global frame feature $F_I$ reveals its vital role to DGT.

\setlength{\tabcolsep}{1pt}
\begin{wraptable}[8]{r}{0.5\textwidth}
\vspace{-10pt}
    \small
    \centering
    \caption{Study of model components}
    \label{tab:aba}
    \vspace{4pt}
    \scalebox{0.7}{
    \begin{tabular}{l|cc|ccc|c}
    \multirow{2}*{Models} & \multicolumn{2}{c|}{TGIF-QA}& \multicolumn{4}{c}{NExT-QA Val}  \\ \cline{2-7}
    ~ & Action & Trans & Acc@C & Acc@T & Acc@D & Acc@All \\ 
    \hline
     VGT & {\bf95.0} & {\bf97.6} & {\bf52.28} & {\bf55.09} & 64.09 & {\bf55.02}  \\ \hline
     w/o DGT & 89.6 & 95.4 &  50.10 & 52.85 & 64.48 & 53.22 \\
     w/o TTrans &  94.0 & 97.6 & 50.86 & 53.04 & 64.86 & 53.74   \\
     w/o NTrans & 94.5 & 97.4 & 50.79 & 54.22 & 63.32 & 53.84 \\ 
     w/o ETrans & 94.8 & 97.4 & 51.25 & 54.34 & 64.48 & 54.30  \\
     w/o $F_I$ & 93.5 & 97.0 & 50.44 & 53.97 & 63.32 & 53.58 \\ 
    \hline
    Comp$\rightarrow$\text{CLS} & 70.1 & 79.9 & 42.96 & 46.96 & 53.02 & 45.82 \\ 
    \end{tabular}
    }
    \vspace{-0.3cm}
\end{wraptable}

\textbf{Similarity Comparison \vs Classification.}
We study a model variant by concatenating the outputs of the DGT module with the token representations from BERT in a way analogous to ClipBERT \cite{lei2021less}. The formed text-video representation sequence is fed to a cross-modal transformer for information fusion. Then, the output of the `[CLS]' token is fed to a $|\mathcal{A}|$-way classifier in open-ended QA or a $1$-way classifier for binary relevance in multi-choice QA following \cite{jang2017tgif,le2020hierarchical,xiao2021video}. As can be seen from the bottom part of Table~\ref{tab:aba}, this classification model variant (Comp $\rightarrow$ CLS) leads to drastic performance drops. To be complete, we also conduct additional experiments on the FrameQA task which is set as open-ended QA. Again, we find that the accuracy drops from 61.6\% to 56.9\%. A detailed analysis of the performances on the training and validation sets (see Appendix \ref{app:cls}) reveals that the CLS-model suffers from serious over-fitting on the target datasets. The experiment demonstrates the superiority of solving QA by relevance comparison instead of answer classification.

\begin{wrapfigure}[8]{r}{0.5\textwidth}
 \vspace{-32pt}
  \begin{center}
    \includegraphics[width=.5\textwidth]{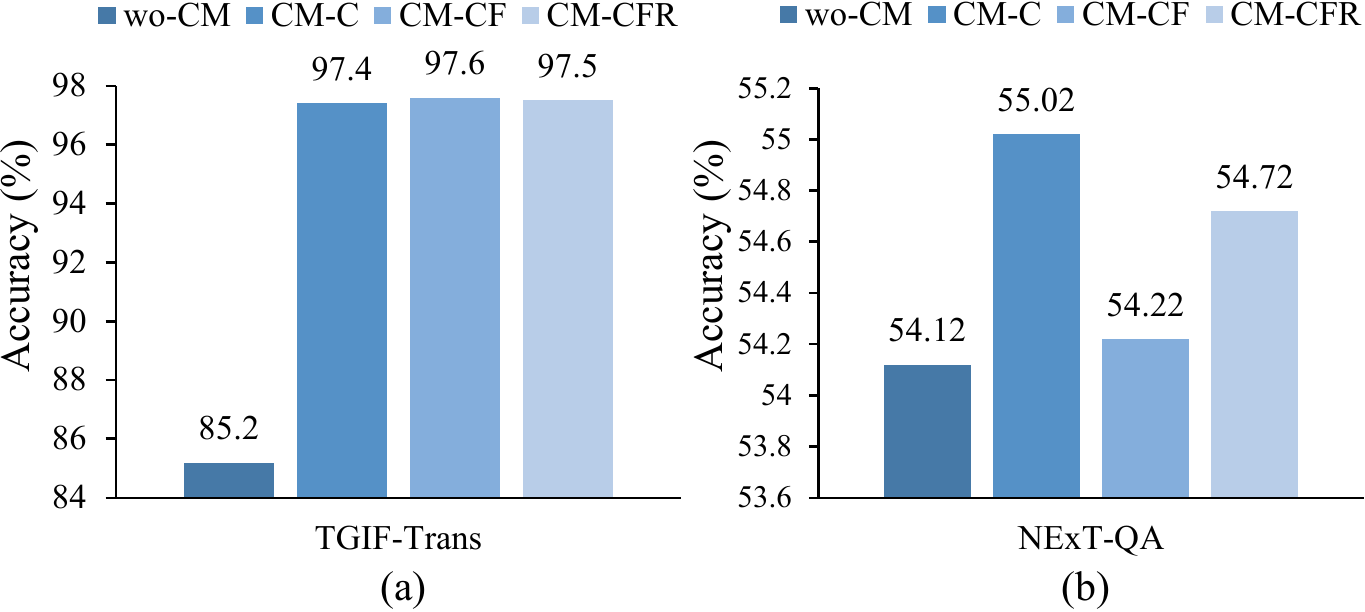}
  \end{center}
   \vspace{-23pt}
  \caption{Study of Cross-modal Interaction.}
  \label{fig:cm}
\end{wrapfigure}

\textbf{Cross-modal Interaction.}
Fig.~\ref{fig:cm} investigates several implementation variants of the cross-modal interaction module as depicted in Sec.~\ref{sec:cm}. The results
suggest that it is better to integrate textual information at both the frame- and clip-level outputs (CM-CF) for TGIF-QA, while our default interaction at the clip-level outputs (CM-C) brings the optimal results on NExT-QA. 
Compared with the baselines that do not use cross-modal interaction, all three kinds of interactions improve the performances. We notice that the cross-modal interaction improves the accuracy on TGIF-QA by more than 10\%. A possible reason is that the GIFs are trimmed short videos that only contain the QA-related visual contents. This greatly eases the challenge in spatial-temporal grounding of the positive answers, especially when most of the negative answers are not presence in the short GIFs. Thus, the cross-modal interaction performs more effectively on this dataset. The videos in NExT-QA are not trimmed, thereby the improvements are relatively smaller.
Base on these observations, we perform cross-modal interaction at both the frame- and clip-level outputs for the temporal reasoning tasks in TGIF-QA, and keep the default implementation for other datasets.

\subsection{Pretraining and Finetuning}
\label{sec:ptft}
\begin{figure*}[t!]
  \begin{center}
    \includegraphics[width=1.0\textwidth]{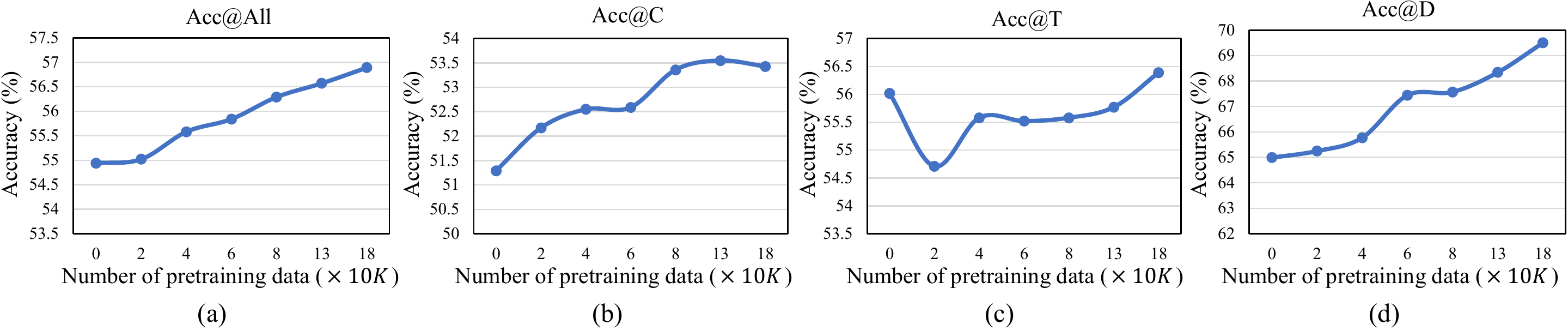}
  \end{center}
  \vspace{-0.6cm}
  \caption{Results of pretraining with different amounts of data.}
  \label{fig:ptp}
  \vspace{-0.2cm}
\end{figure*}

Table~\ref{tab:pt} presents a comparison between VGT with and without pretraining. We can see that pretraining can steadily boost the QA performance, especially on NExT-QA. The relatively smaller improvements on TGIF-QA could be due to that TGIF-QA dataset is large, and has enough annotated data for fine-tuning. As such, pretraining helps little \cite{zoph2020rethinking}. Besides, we find that finetuning with masked language modelling (MLM) can improve the generalization from val to test set, and thus achieves the best overall accuracy (\ie~55.7\%) on NExT-QA test set. Fig.~\ref{fig:ptp} studies the QA performances on NExT-QA val set with respect to different amounts of pretraining data. Generally, there is a clear tendency of performance improvements for the overall accuracy (Acc@All) when more data is available. A more detailed analysis shows that these improvements mostly come from a stronger performance in answering causal (Acc@C) and descriptive (Acc@D) questions. For temporal questions, it seems that pretraining with more data does not help much. Therefore, 
to boost performance, it is promising to add more data or explore a better way to handle temporal languages. 

\setlength{\tabcolsep}{1pt}
\begin{table*}[t!]
    \small
    \centering
    \caption{Study of cross-model pretraining. Results on NExT-QA are with 20 regions.
    }
    \label{tab:pt}
    \vspace{-0.2cm}
    \scalebox{0.78}{
        \begin{tabular}{l|cc|ccc|c|ccc|c}
        \multirow{2}*{Methods} & \multicolumn{2}{c|}{TGIF-QA}& \multicolumn{4}{c|}{NExT-QA Val} & \multicolumn{4}{c}{NExT-QA Test} \\ \cline{2-11}
        ~ & Action$\dagger$ & Trans$\dagger$ & Acc@C & Acc@T & Acc@D & Acc@All & Acc@C & Acc@T & Acc@D & Acc@All \\ 
        \hline
        VGT & 59.9 & 70.5 & 51.29 & 56.02 & 64.99 & 54.94 & 50.82 & 52.29 & 63.27 & 53.51 \\ \hline
        VGT (FT w/ QA) & 60.2 & 71.0 & {\bf53.93} & 56.20 & {\bf70.14} & {\bf57.19} & 51.73 & 53.78 & 67.05 & 54.88 \\
        VGT (FT w/ QA \& MLM) & {\bf60.5} & {\bf71.5} & 53.43 & {\bf56.39} & 69.50 & 56.89 & {\bf52.78} & {\bf54.54} & {\bf67.26} & {\bf55.70} \\ 
        \end{tabular}
    }
    \vspace{-0.5cm}
\end{table*}


\subsection{Qualitative Analysis}
\begin{figure*}[t!]
  \begin{center}
    \includegraphics[width=1.0\textwidth]{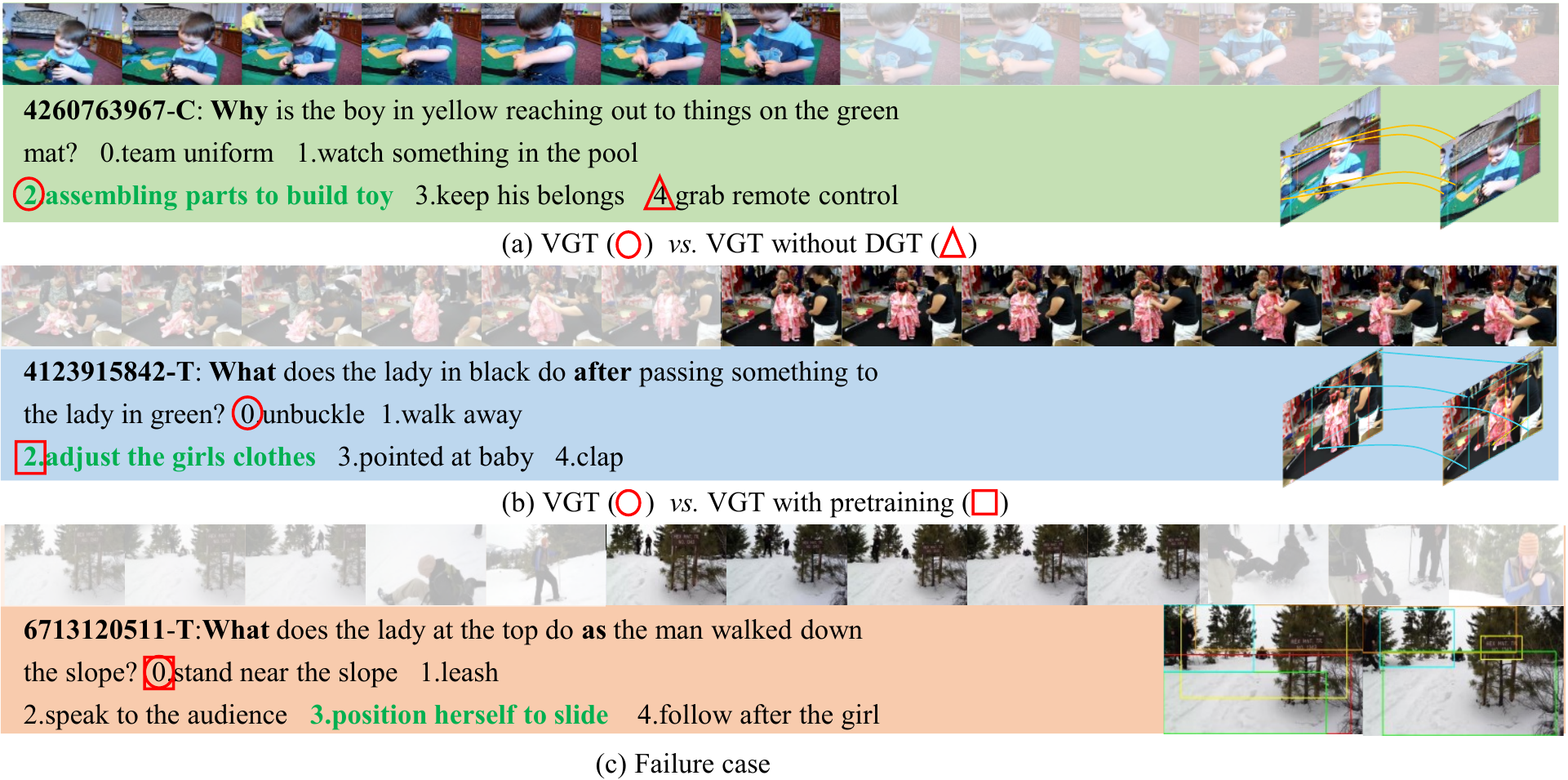}
  \end{center}
  \vspace{-0.7cm}
  \caption{Result visualization on NExT-QA \cite{xiao2021next}. The ground-truth answers are in green. 
  }
  \label{fig:vis}
  \vspace{-0.5cm}
\end{figure*}
In Fig.~\ref{fig:vis}, we qualitatively analyze the benefits of both dynamic graph transformer and pretraining. The example in (a) shows that the model without the DGT module is prone to predicting atomic or contact actions (\eg~`grab') that can be captured at static frame-level. (b) shows that the model without pretraining fails to predict the answer that is highly abstract (\eg~`adjust'). Finally, we show a failure case in (c). It indicates that our model tends to predict distractor answers that are semantically close to the questions when the object of interests in the video are small and the detector fails to detect it. Keeping more detected regions could be helpful, but one needs to carefully balance the graph complexity as well as the inference efficiency. Another alternative is to perform modulated detection as in \cite{kamath2021mdetr}, we leave it for future exploration.
\section{Conclusions}
We presented video graph transformer which explicitly exploits the objects, their relations, and dynamics, to improve visual reasoning and alleviate the data-hungry issue for VideoQA. Our extensive experiments show that VGT can achieve superior performances as compared with previous SoTA methods on tasks that challenge temporal dynamic reasoning. The performance even surpasses those methods that are pretrained on large-scale vision-text data. To study the learning capacity of VGT, we further explored pretraining on weakly-paired video-text data and obtained promising results. With careful and comprehensive analyses of the model, we hope this work can encourage more efforts in designing effectiveness models to alleviate the burden of handling large-scale data, and also promote VQA research that goes beyond a holistic recognition/description to reason about the fine-grained video details.

\section*{\small{Acknowledgements}}
\vspace{-0.3cm}
\small{This research is supported by the Sea-NExT joint Lab. Major work was done when Junbin was a research intern at Sea AI Lab. We greatly thank Angela Yao as well as the anonymous reviewers for their thoughtful comments towards a better work.}
%
%
\bibliographystyle{splncs04}
\bibliography{egbib}

\clearpage
\appendix
\input{append/append.tex}

\end{document}

%% file: append/append.tex
\section{Data Statistics}
\label{app:dset}
The statistical details of the experimented datasets are presented in Table~\ref{tab:dataset}. For better comparison with previous works, we focus on the multi-choice QA task in NExT-QA \cite{xiao2021next} though it has also defined open-ended QA. For TGIF-QA \cite{jang2017tgif}, we also conduct experiments on a latest version \cite{peng2021progressive} which generates more challenging negative answers for each question in the multi-choice tasks. In particular, we further fix the `redundant answer' issue as we find that there are about 10\% of questions have redundant candidate answers and some of the candidate answers are even identical to the correct one. The rectified annotations will be released along with the code.

\setlength{\tabcolsep}{2.3pt}
\begin{table*}[t!]
\small
\centering
\caption{Data statistics. OE: Open-Ended QA. MC: Multi-Choice QA, VLen (s): Average video length in seconds.}
\vspace{-0.8em}
\begin{threeparttable}
    \scalebox{0.7}{
    \begin{tabular}{l|lccccc|c}
        Datasets & Main Challenges & \#Videos/\#QAs & Train & Val & Test & VLen (s) & QA \cr
        \hline
        NExT-QA \cite{xiao2021next} & Causal \& Temporal Interaction  & 5.4K/48K & 3.8K/34K & 0.6K/5K & 1K/9K & 44 & MC \cr
         \cline{1-8}
         \multirow{3}*{TGIF-QA \cite{jang2017tgif}} 
         & Repetition Action  & 22.8K/22.7K & 20.5K/20.5K & -& 2.3K/2.3K & 3 & MC\cr
         & State Transition  & 29.5K/58.9K & 26.4K/52.7K & -& 3.1K/6.2K & 3 & MC\cr
         & Frame QA         & 39.5K/53.1K & 32.3K/39.4K & -& 7.1K/13.7K & 3 & OE\cr
         \cline{1-8}
         MSRVTT-QA \cite{xu2017video} & Descriptive QA  & 10K/ 244K & 6.5K/159K & 0.5K/12K & 3K/73K & 15& OE\cr
    \end{tabular}
    }
    \vspace{-0.5cm}
\label{tab:dataset}
\end{threeparttable}
\end{table*}

\section{Implementation Details}
\label{app:imp}
For training with QA annotations, we firstly train the whole model (except for the object detection model) end-to-end, and then freeze BERT to fine-tune the other parts of the best model obtained at the $1$st stage. The best results in the two stages are determined as final results. Note that our hyper-parameters are mostly searched on the NExT-QA validation set and kept unchanged for other datasets. The maximum epoch varies from 10 to 30 among different datasets. For pretraining with data crawled from the Web, we randomly select 0.18M video-text data (less than 10\%) from WebVid2.5M
\footnote{\url{https://m-bain.github.io/webvid-dataset/}}
\cite{bain2021frozen}. The videos are then extracted at 5 frames per second and are processed in the same way as for QA. We then optimize the model with an initial learning rate of $5\times10^{-5}$ and batch size 64. The number of negative descriptions of a video for cross-modal matching is set to 63, and they are randomly selected from the descriptions of other videos in the whole training set. Besides, a text token is corrupted at a probability of 15\% in masked language modelling. Following \cite{yang2021just}, a corrupted token will be replaced with 1) the `[MASK]' token by a chance of 80\%, 2) a random token by a chance of 10\%, and 3) the same token by a chance of 10\%. We train the model by maximal 2 epochs which gives to the best generalization results, and it takes about 2 hours.

\section{Additional Model Analysis}
\label{app:addexp}
\subsection{Similarity Comparison \vs Classification} 
\label{app:cls}
\begin{figure*}[t!]
  \begin{center}
    \includegraphics[width=0.7\textwidth]{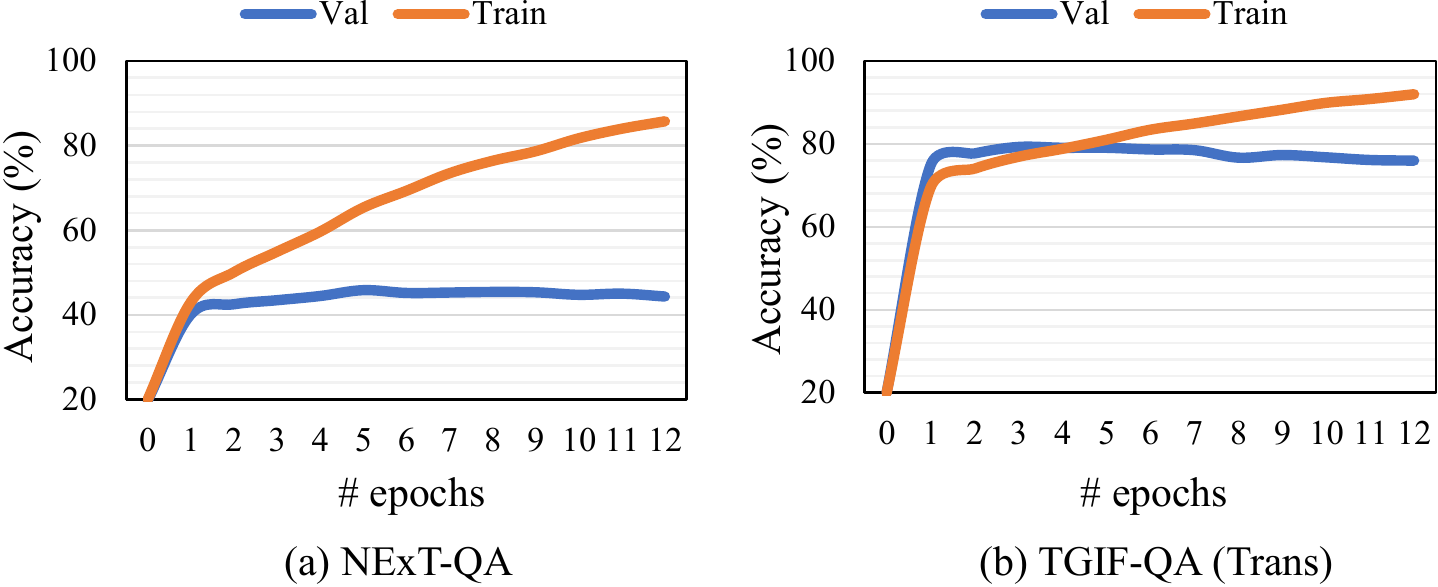}
  \end{center}
  \vspace{-0.4cm}
  \caption{Accuracy with regard to different training epochs.}
  \label{fig:of}
\end{figure*}

To study the reason for the poor performance of the classification model variant described in Sec.~\ref{sec:analysis} of the main text, we visualize the training and validation accuracy with regard to different training epochs in Fig.~\ref{fig:of}. The results indicate that the classification model variant suffers from serious over-fitting issues, especially on NExT-QA \cite{xiao2021next} whose QA contents are relative complex but with less training data. To study whether the problem comes from the classification formulation or the cross-modal transformer, we further substitute the cross-modal transformer (CM-Trans) with our cross-modal interaction (CM) module introduced in Sec.~\ref{sec:cm} of the main text. We find that such a substitution can slightly alleviate the problem. For example, on NExT-QA val set, the accuracy increases from 45.82\% to 46.98\%. Nevertheless, the performance is still much worse than a comparison-based model implementation (\ie~55.02\%). This experiment reveals two facts: 1) Formulating QA problem as classification is the major cause for the weak performance. 2) The cross-modal transformer exacerbates the over-fitting problem, possibly because it involves additional parameters.

\subsection{Study of Video Sampling}
\label{app:sap}
\begin{figure}[t!]
  \begin{center}
    \includegraphics[width=0.6\textwidth]{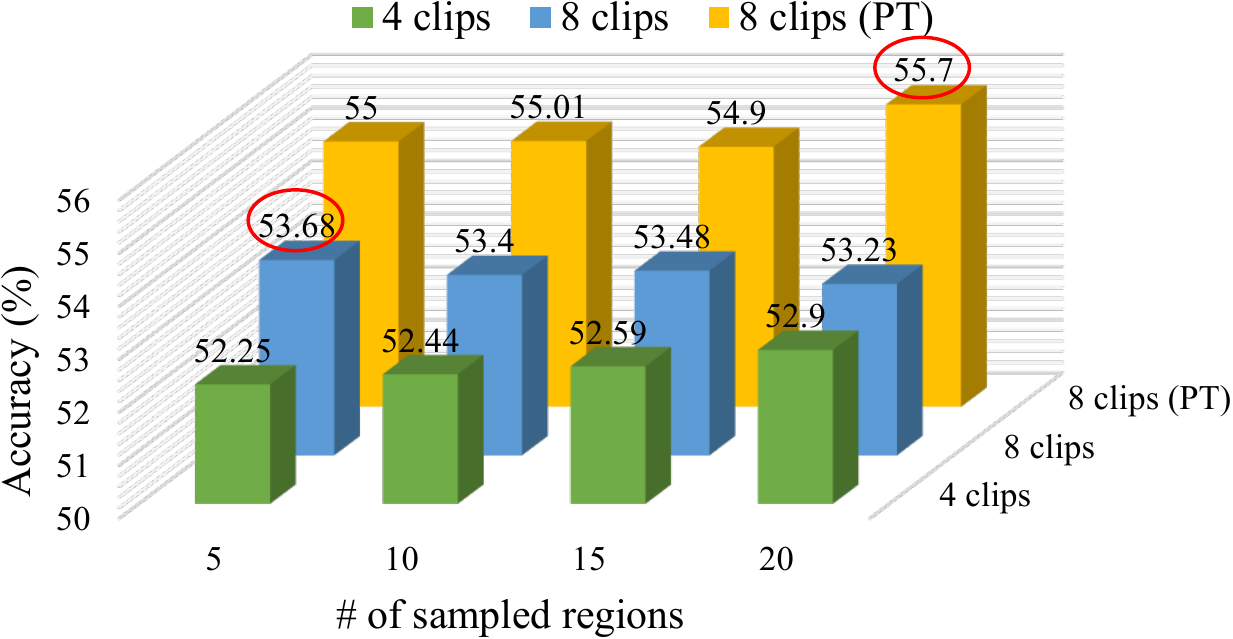}
  \end{center}
  \vspace{-0.2cm}
  \caption{Investigation of sampled video clips and region proposals per frame. Results are reported on NExT-QA test set.}
  \label{fig:region}
\end{figure}

In Fig.~\ref{fig:region}, we study the effect of sampled video clips and region proposals on NExT-QA \cite{xiao2021next} test set. Regarding the number of sampled video clips, we find that the setting of 8 clips steadily wins on 4 clips. This is understandable as the videos in NExT-QA are relatively long. As for the sampled regions, when learning the model from scratch, the setting of 5 regions gives relatively better result, \eg, 53.68\%. Nonetheless, when pretraining are considered, the setting of 20 regions gives better result, \eg, 55.70\%. Such difference could be due to that learning with more regions can yield over-fitting issues when the dataset is not large enough, since the constructed graph become much larger and more complex. Our speculation is also supported by the fact that the accuracy increases with the number of sampled regions when we only sample 4 video clips and thus less number of total graph nodes.

\subsection{Model Efficiency}
\label{app:eff}
\setlength{\tabcolsep}{3pt}
\begin{table}[t!]
    \small
    \centering
    \caption{Comparison of memory and time based on NExT-QA \cite{xiao2021next}. (2m$\times$8: 2 minutes per epoch and 8 epochs in total.) 
    }
    \vspace{-0.2cm}
    \scalebox{0.8}{
    \begin{tabular}{l|c|c|cc|cc}
    \multirow{2}*{Models} & \multirow{2}*{Acc@All} & \multirow{2}*{\#Params (M)} & \multicolumn{2}{c|}{GPU Memory} & \multicolumn{2}{c}{Time} \\ 
    \cline{4-7}
    ~ & ~ & ~ & Train & Infer & Train & Infer(FLOPs) \\ 
    \hline
    VQA-T [52] & 45.30 & 156.5 & 5.6G & 2.6G & 2m$\times$8 & 2448M \\ 
    VGT (BERT) & 55.02 & 133.7 & 16.2G & 3.9G & 7m$\times$5 & 7121M \\ 
    VGT (DistilBERT) & 53.46 & 90.5 & 10.0G & 3.5G & 5m$\times$7 & 3922M \\ 
    \end{tabular}
    }
    \label{tab:tm}
    \vspace{-0.4cm}
\end{table}

We compare VGT with VQA-T \cite{yang2021just} in Tab.~\ref{tab:tm} for better understanding of the memory and time cost. Experiments are done on 1 Tesla V100 GPU with batch size 64. We use 1 example to report inference FLOPs.
\textbf{{Memory:}} VGT has less training parameters (133.7M \vs 156.5M) and thus smaller model size than VQA-T (511M \vs 600M). The BERT encoder in VGT takes 82\% of the parameters, the vision part is lightweight with only 24M parameters. VGT needs more GPU memory for training. Yet, the memory for inference are fairly small and close to that of VQA-T. We also implement a smaller version of VGT by replacing BERT with DistilBERT \cite{sanh2019distilbert} as in VQA-T. With nearly 0.6$\times$ number of VQA-T's parameters (90.5/156.5M), we can still achieve strong performances (\ie~53.46\%).
\textbf{{Time:}} Our FLOPs on 1 example is $\sim$2.9$\times$ that of VQA-T and $\sim$1.6$\times$ if we use DistilBERT. However, VGT converges much faster and needs much fewer epochs (total FLOPs) to get results superior to VQA-T when training with the same data. For example, on NExT-QA, VGT's result at epoch 2 (50.16\%) already significantly surpasses VQA-T's best result (45.30\%) achieved at epoch 8. Also, VGT's result without pretraining can surpasses that of VQA-T pretrained with million-scale data. In this sense, VGT needs much fewer total FLOPs than VQA-T and other similar pretrained models for visual reasoning. 